\documentclass[10pt, a4paper]{article}

\usepackage{lrec-coling2024}
\usepackage{todonotes}

\usepackage{natbib}
\usepackage{multibib}
\makeatletter
\def\@mb@citenamelist{cite,citep,citet,citealp,citealt,citepalias,citetalias}
\makeatother
\newcites{languageresource}{~}

\usepackage{graphicx}
\usepackage{tabularx}
\usepackage{soul}

\usepackage{xcolor}
\usepackage{hyperref}
 \definecolor{darkblue}{rgb}{0, 0, 0.5}
  \hypersetup{colorlinks=true, citecolor=darkblue, linkcolor=darkblue, urlcolor=darkblue}

\usepackage{xstring}

\usepackage{color}
\usepackage{multirow}
\usepackage{svg}
\usepackage{authblk}
\usepackage{caption}
\usepackage{placeins}

\usepackage{babel}

\title{TeClass: A Human-Annotated Relevance-based Headline Classification and Generation Dataset for Telugu}

\name{Gopichand Kanumolu\textsuperscript{*}\thanks{*Authors contributed equally},
  Lokesh Madasu\textsuperscript{*}\footnotemark[1],
  Nirmal Surange,
  Manish Shrivastava}

\address{Language Technologies Research Center, KCIS, IIIT Hyderabad, India.\\
\texttt{\{gopichand.kanumolu, lokesh.madasu, nirmal.surange\}@research.iiit.ac.in} \\
\texttt{m.shrivastava@iiit.ac.in}
}

\abstract{
News headline generation is a crucial task in increasing productivity for both the readers and producers of news. This task can easily be aided by automated News headline-generation models. However, the presence of irrelevant headlines in scraped news articles results in sub-optimal performance of generation models. We propose that relevance-based headline classification can greatly aid the task of generating relevant headlines.
Relevance-based headline classification involves categorizing news headlines based on their relevance to the corresponding news articles.
While this task is well-established in English, it remains under-explored in low-resource languages like Telugu due to a lack of annotated data.
To address this gap, we present TeClass, the first-ever human-annotated Telugu news headline classification dataset, containing 78,534 annotations across 26,178 article-headline pairs.
We experiment with various baseline models and provide a comprehensive analysis of their results. 
We further demonstrate the impact of this work by fine-tuning various headline generation models using TeClass dataset. The headlines generated by the models fine-tuned on highly relevant article-headline pairs, showed about a 5 point increment in the ROUGE-L scores.
To encourage future research, the annotated dataset as well as the annotation guidelines will be made publicly available. 
 \\ \newline  }

\begin{document}

\maketitleabstract

\section{Introduction}

A headline is a single-sentence summary of a news article that aspires to present a concise and factual account of the story described in the article. 
It is a crucial element in drawing the reader's attention to the article's content and is designed to engage the reader. Headlines are often the only thing that the reader sees before deciding whether to click and read further. They act as a filter, allowing the reader to quickly decide if the story is relevant or interesting to them. In today's rapidly evolving information landscape, the task of assessing the relationship between news headlines and their corresponding articles has become a critical challenge, and this task can be conceptualized in various forms such as fake news detection, misinformation detection, incongruent news headline detection, headline classification, etc.\\

Generation of a relevant headline can be a challenging and time-consuming task. In most cases, barring sensational and click-bait headlines, the headline needs to draw out the most relevant aspects of the article in a single meaningful string\footnote{Headline need not be a complete sentence}. Therefore, headline generation is often posed as a summarization task \cite{rush,headline2020, Bukhtiyarov}.  But, despite the existence of multiple article-headline datasets, the generation of relevant headlines remains a challenge, especially for low-resource languages. This can be attributed to the noise present in the datasets in the form of irrelevant headlines \cite{jin2020hooks}. 

The relevance or irrelevance of a headline with respect to the article has been explored by \citet{FNCDataset} in the Fake News Challenge (FNC-1) to determine the stance of a news article relative to the headline. FNC-1 dataset is an extension of the work of \citet{Emergent}. The FNC-1 dataset contains 49,972 article-headline pairs labeled with one of the four categories namely Agrees, Disagrees, Discusses, and Unrelated. However, it is important to note that the Unrelated category, constituting 73\% of the dataset is generated by pairing the headlines and articles belonging to different topics at random, and hence may not reflect the original relation between article and headline \cite{headline_incongruency}.


\begin{figure}[t]
    \includegraphics[width=0.5\textwidth]{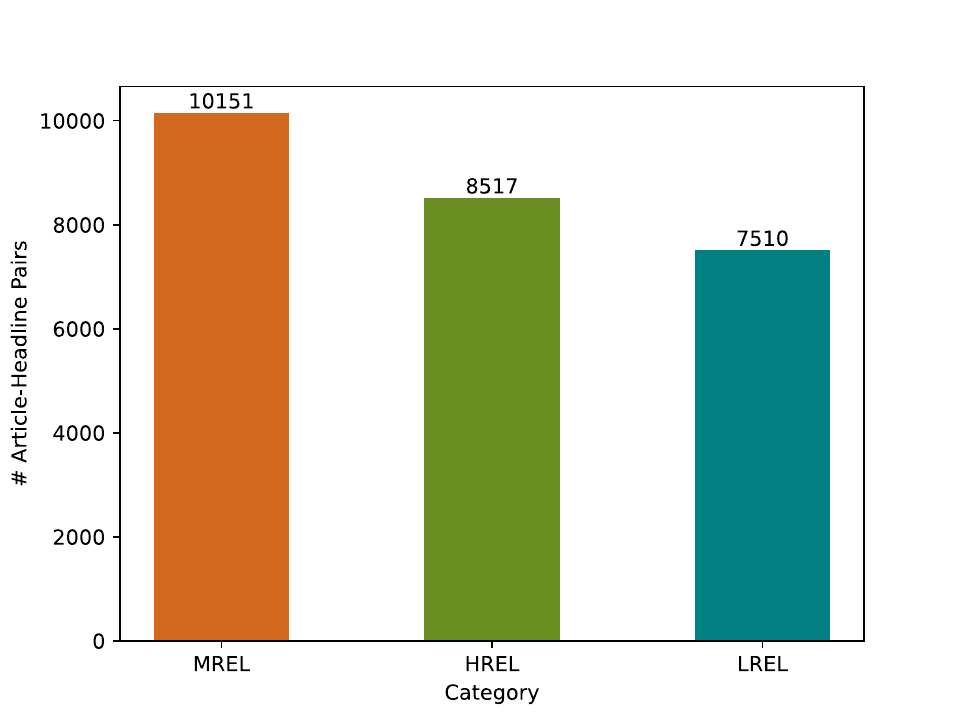}
    \caption{Category distribution in TeClass. HREL: Highly Related, MREL: Moderately Related, LREL: Least Related}
    \label{fig:stance_category_bar_plot}
\end{figure}

We believe that the generation of relevant headlines is contingent on the quality of the data presented, especially for low-resource languages like Telugu. We have observed that for low resource languages like Telugu, the ratio of highly relevant headlines versus not-so-relevant or irrelevant headlines is badly skewed towards irrelevance (Figure \ref{fig:stance_category_bar_plot}). This might be due to market pressures for publication houses to draw customers to click-baits or might also be due to the cognitively challenging nature of headline creation task. The impact of this imbalance is seen in wasted time for viewers. Automatic headline generation might help in the latter case but the skew in the distribution of informative headlines means that most of the training compute for the models is spent training on non-informative/irrelevant headlines, eventually impacting the performance negatively. Therefore, we propose that headline generation models should only be trained on highly related article-headline pairs. This requires a pre-processing step of headline relevance classification.

With this motivation, we have created a novel dataset for relevance-based headline classification that reflects the nuances of the real-world news article-headline pairs in the Telugu language. Our key contributions in this paper are summarized as follows:
\begin{enumerate}
    \item We present "TeClass", a large, diverse, and high-quality human-annotated dataset for a low-resource language Telugu, containing 26,178 article-headline pairs annotated for headline classification with one of the three categories: 
    \begin{itemize}
        \item Highly Related (HREL): The headline is highly related to the article.
        \item Moderately Related (MREL): The headline is moderately related to the article.
        \item Least Related (LREL): The headline is vaguely related to the article.
    \end{itemize}
    \item We present a comprehensive analysis of various baseline models employed for headline classification on this dataset. 
    \item We present baseline headline generation models to demonstrate that the task of relevant headline generation is best served when the generation models are trained on high-quality relevant data even if the available relevant article-headline pairs are significantly less in number.
\end{enumerate}
To lay the foundation for future work, our dataset and models are made publicly 
available\footnote{\url{https://github.com/ltrc/TeClass}}.\\

\section{Dataset}
\label{sec:dataset}
\subsection{Selecting the Article-Headline Pairs for Annotation}

We collect the news article-headline pairs from multiple news websites using web scraping. As websites often follow their own style of writing the news, to mitigate any potential bias towards a particular style of news reporting, we gathered data from a diverse range of news websites. These websites covered a broad spectrum of domains, including State, National, International, Entertainment, Sports, Business, Politics, Crime, and COVID-19.

However, web scraping from multiple sources posed a significant challenge due to the dynamic nature of websites. Each website has its unique structure, necessitating a thorough understanding of its individual layouts to ensure the extraction of data without loss of information or the introduction of extraneous noise. To address this challenge, we developed custom site-specific web scrapers tailored to each news website. These scrapers were designed to extract three essential components: the text of the news article, the headline, and the name of the news domain. Our extraction methodology was carefully crafted to exclude any undesirable elements, such as advertisements, URLs pointing to related articles, and embedded social media content within the news body.\\

\subsection{Annotation }

The relationship between a news headline and its corresponding article can occur in many ways. In ideal cases, the headline summarizes the core idea of the article. Some headlines are designed to capture attention and generate clicks, often by using provocative or sensational language. In some instances, headlines can be misleading, either intentionally or unintentionally, by not accurately representing the information presented in the article. Occasionally, headlines may focus on less important details of the article.\\

We employed crowd-sourcing for the annotation process, engaging native Telugu-speaking volunteers.  We presented the following instructions to the annotators, and the annotators were asked to assign one of the three primary categories: High relevance (HREL), Medium relevance (MREL), and Low relevance (LREL) after reading the headline and its corresponding article. They are also instructed to assign a secondary sub-class for each article.\\

\begin{figure*}[]
    \centering
        \includegraphics[clip, trim=2cm 2cm 2cm 2cm, width=1\textwidth]{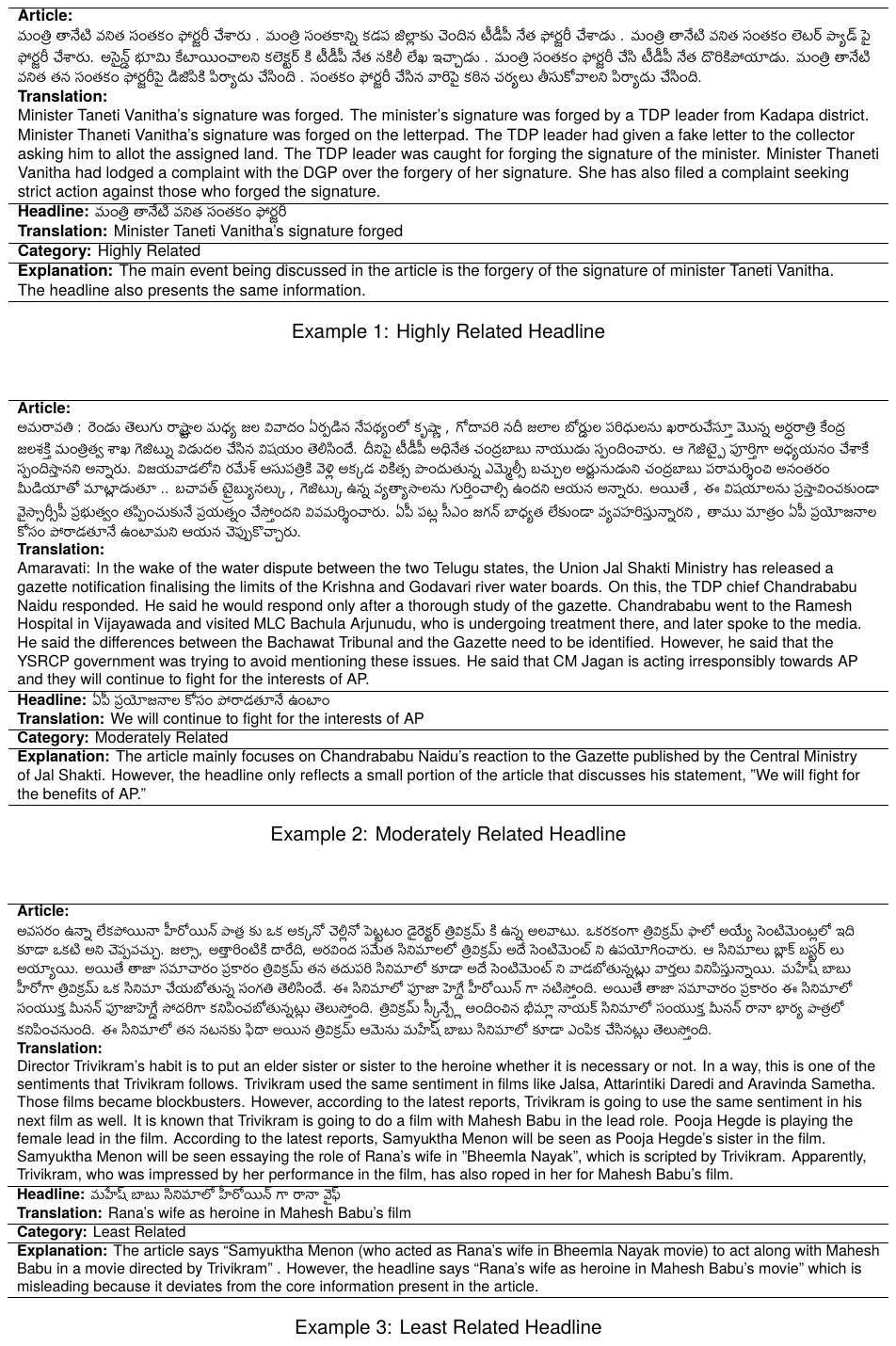}
    \vspace*{-7mm}
    \caption{Examples of relevance-based headline classification for each category}
    \label{fig:TeluguHCExamples}
    \vspace*{-1mm}
\end{figure*}

\textbf{HREL}: The headline is highly related to the article content if it satisfies the following condition (Example 1 of Figure \ref{fig:TeluguHCExamples}):
\begin{itemize}
    \item Factual Main Event (FME): The headline is mostly explicitly present in the article and represents the main event addressed in the article which is factually correct. 
\end{itemize}

\textbf{MREL}: The headline is moderately related to the article content if it satisfies any of the following conditions (Example 2 of Figure \ref{fig:TeluguHCExamples}):
\begin{itemize}
    \item  Strong Conclusion (STC): The headline is not explicitly present (in the same words) in the article, but it can be inferred from the article and represents the majority of the article content.
    \item Factual Secondary Event (FSE): The headline represents a secondary event addressed in the article which is factually correct.
    \item Weak Conclusion (WKC):  The headline is not explicitly present (in the same words) in the article, and it has been inferred from only a small portion of the article content.
\end{itemize}

\textbf{LREL}: The headline is least related to the article content if it satisfies any of the following conditions (Example 3 of Figure \ref{fig:TeluguHCExamples}):
\begin{itemize}
    \item Sensational (SEN): The Headline is intended to catch the attention of the reader, by reporting biased/emotionally loaded impressions/controversial statements that manipulate the truth of the story.
    \item Clickbait (CBT): A headline that tempts the reader to click on the link, where there is an extreme disconnect between what is being presented on the front side of the link (headline) versus what is on the click-through side of the link (article). 
    \item Misleading Conclusion (MLC): A headline that vaguely draws a conclusion about the article that is not supported by the facts in the article.
    \item Unsupported Opinion (USO): A headline that is an opinion about an article's event/subject but is not supported by the article.
\end{itemize}


\FloatBarrier

A pilot study involving a small-scale trial annotation was conducted to ensure that the annotation guidelines were clear and unambiguous. We explained the guidelines to the annotators to ensure that the annotators understood the task's objectives. Additionally, we closely monitor the annotation process and conduct query resolution sessions to provide assistance in handling ambiguous, or difficult examples. we assign each article-headline pair to 3 annotators, and the final category for a pair is chosen based on the majority vote among the 3 annotations.

\subsection{Annotated Dataset Statistics}


\begin{figure}[t]
    \raggedright
    \includegraphics[width=0.5\textwidth]{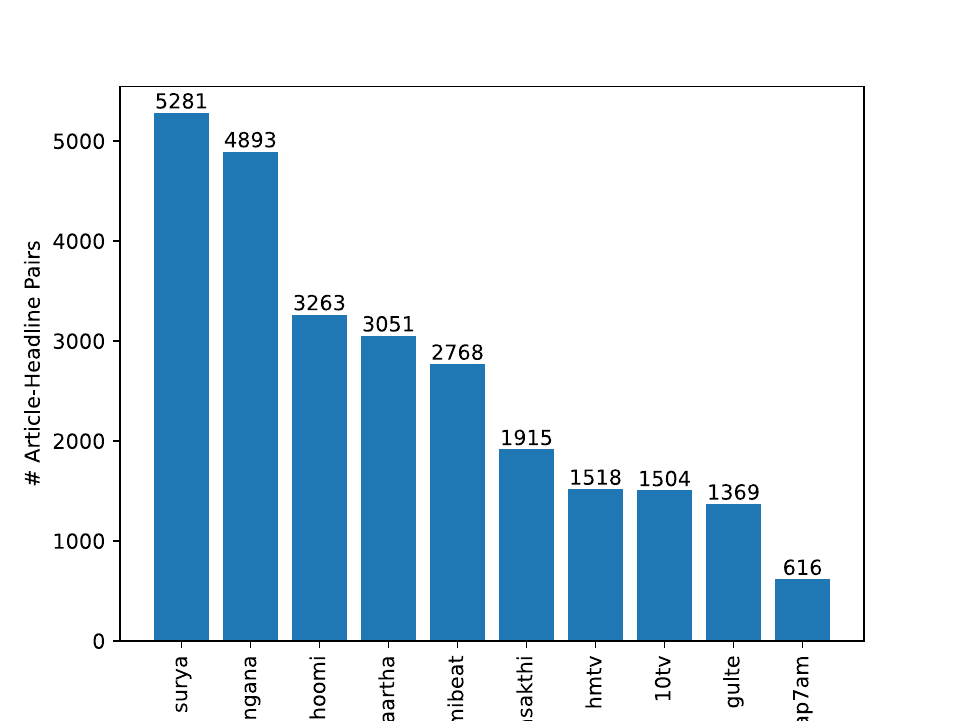}
    \caption{News website distribution in TeClass}
    \label{fig:website-dist}
\end{figure}

\begin{figure}[t]
    \centering
    \includegraphics[width=0.5\textwidth]{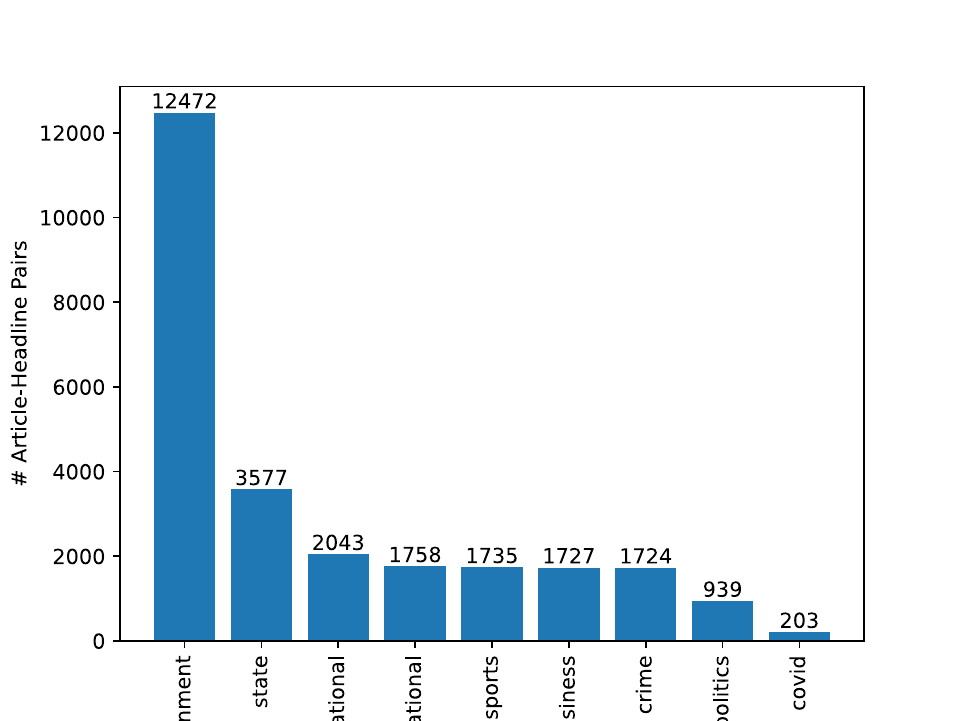}
    \caption{News domain distribution in TeClass}
    \label{fig:domain-dist}
\end{figure}

In this section, we present the statistics of the annotated dataset. Since each article-headline pair is annotated by 3 annotators, we get a total of 78,534 annotations for 26,178 unique article-headline pairs. The category-wise counts of the dataset are presented in Figure \ref{fig:stance_category_bar_plot}. As mentioned earlier, the dataset contains article-headline pairs from multiple websites with a diverse set of news domains, the website-wise and 
domain-wise pairs distribution is detailed in 
Figure \ref{fig:website-dist}, and 
Figure \ref{fig:domain-dist} respectively. \\

\textbf{Data Splits:} 
We allocated 70\% for training, 15\% for development and 15\% for testing.
To ensure unbiased performance and prevent category bias, we applied stratified sampling techniques. This ensures even distribution of articles from all 3 categories across the training, development, and test sets. The category-wise counts in each data split are presented in Table \ref{tab:cat_wise_counts_in_splits}. Further statistical details of the TeClass dataset are available in Table \ref{tab:TeClass_Statistics}.

\begin{table}[h]
\centering
\begin{tabular}{c|c|c|c}
\hline
 &  \textbf{Train} & \textbf{Dev} & \textbf{Test} \\ \hline
HREL              & 5962           & 1277         & 1278          \\ 
MREL              & 7105           & 1523         & 1523          \\ 
LREL              & 5257           & 1127         & 1126          \\ \hline
\end{tabular}
\caption{Category-wise counts in each data split}
\label{tab:cat_wise_counts_in_splits}
\end{table}

\paragraph{Inter-Annotator Agreement:}
Having multiple annotators (typically three or more) for annotation tasks is vital for several reasons. They enable the measurement of inter-annotator agreement, helping to identify and address ambiguous or challenging cases. Multiple annotators also help mitigate individual bias and promote a balanced, objective annotation process ensuring the robustness and quality of the annotated dataset. We use Fleiss' Kappa metric proposed by \citet{randolph2005free} and it resulted in an encouragingly high score of 0.77, indicating a substantial agreement among the annotators. 

\begin{table*}[t]
\centering
\begin{tabular}{c|c|c|c}
\hline
\textbf{}                     & \textbf{Train} & \textbf{Dev} & \textbf{Test} \\ \hline
Article-Headline pairs        & 18,324          & 3,927         & 3,927          \\ 
Average sentences in article  & 10.30          & 10.25        & 10.29         \\ 
Average sentences in headline & 1.06           & 1.06         & 1.05          \\ 
Average tokens in article     & 126.33         & 126.70       & 126.39        \\ 
Average tokens in headline    & 6.16           & 6.15         & 6.11          \\ 
Unique tokens in articles     & 204959         & 76279        & 76070             \\ 
Unique tokens in headlines    & 28785          & 9894         & 10008             \\ \hline
Average LEAD-1 score          & 16.88          & 17.09        & 16.88         \\ 
Average EXT-ORACLE score      & 29.47          & 29.01        & 29.49         \\ \hline
\end{tabular}

\caption{TeClass Statistics}
\label{tab:TeClass_Statistics}
\end{table*}

\section{Headline Classification}
\label{sec:baseline_models}
We experiment with various baseline models including traditional feature-based Machine Learning (ML) models for classification, and also leverage the transfer learning using the state-of-the-art pre-trained BERT \citep{devlin2018bert} models. \\

\textbf{ML baseline models:}
Various participating teams in the FNC-1 challenge make use of features like n-gram overlap, cosine similarity between vector representations of the article, and the headline, and other hand-crafted features \cite{retrospective_analysis_fnc}. We also experiment with various features, and our model architecture is similar to the one proposed by \citet{FNC_baseline} . We use TF-IDF encoding to represent the article, and headline in vector format. To avoid the problem of out-of-vocabulary words, we use subword tokenization that breaks words into smaller subword units, which is vital for morphologically rich languages like Telugu. It resulted in a subword vocabulary of size 2945, which is in turn the dimension of the vector representation of the article, and headline using TF-IDF encoding.  We concatenate the feature vector with the article, and headline representations, and the output of concatenation is passed as input to train the classifier. The feature vector is extracted from the  article-headline pairs using the following methods:
\begin{enumerate}
    \item Cosine similarity: To measure the similarity in content between the article and headline, we compute the cosine similarity between the TF-IDF vector representations of the article and headline.
    \item Novel n-gram percentage: It quantifies the level of uniqueness in a headline by measuring the proportion of n-grams (contiguous sequences of n words) found in the headline but not present in the accompanying article.
    \item LEAD-1: It is the ROUGE-L \citep{XL-Sum_ROUGE}\footnote{\url{https://github.com/csebuetnlp/xl-sum/tree/master/multilingual_rouge_scoring}} score between the headline and the first sentence of the article.
    \item EXT-ORACLE: This score is computed by selecting the sentence from the article that achieves the highest ROUGE-L score with the headline.
\end{enumerate}
We use Logistic Regression (LR), Support Vector Machine (SVM), Multilayer Perceptron (MLP), and Bagging as classification models. All these models use 5-fold cross-validation. We assess model performance using the F1-Score, and the corresponding results are presented in Table \ref{tab:ml_baseline_results}. \\

\begin{table*}[t]
\centering
\begin{tabular}{l|c|ccccc}
\hline
\multirow{2}{*}{\textbf{Feature Vector}} & \multicolumn{1}{l|}{\multirow{2}{*}{\textbf{Classifier}}} & \multicolumn{5}{c}{\textbf{F1 Score}} \\ \cline{3-7} 
 & \multicolumn{1}{l|}{} & \multicolumn{1}{c|}{HREL} & \multicolumn{1}{c|}{MREL} & \multicolumn{1}{c|}{LREL} & \multicolumn{1}{c|}{\begin{tabular}[c]{@{}c@{}}Overall \\ (Weighted)\end{tabular}} & \begin{tabular}[c]{@{}c@{}}Overall \\ (Macro)\end{tabular} \\ \hline
\multirow{4}{*}{Without Feature Vector} & LR & \multicolumn{1}{c|}{0.57} & \multicolumn{1}{c|}{0.50} & \multicolumn{1}{c|}{0.59} & \multicolumn{1}{c|}{0.55} & 0.55 \\ 
 & SVM & \multicolumn{1}{c|}{0.55} & \multicolumn{1}{c|}{0.49} & \multicolumn{1}{c|}{0.57} & \multicolumn{1}{c|}{0.53} & 0.54 \\ 
 & MLP & \multicolumn{1}{c|}{0.55} & \multicolumn{1}{c|}{0.49} & \multicolumn{1}{c|}{0.58} & \multicolumn{1}{c|}{0.54} & 0.54 \\ 
 & Bagging & \multicolumn{1}{c|}{0.55} & \multicolumn{1}{c|}{0.47} & \multicolumn{1}{c|}{0.57} & \multicolumn{1}{c|}{0.52} & 0.53 \\ \hline
\multirow{4}{*}{Cosine Similarity} & LR & \multicolumn{1}{c|}{0.58} & \multicolumn{1}{c|}{0.50} & \multicolumn{1}{c|}{0.59} & \multicolumn{1}{c|}{0.55} & 0.56 \\ 
 & SVM & \multicolumn{1}{c|}{0.56} & \multicolumn{1}{c|}{0.49} & \multicolumn{1}{c|}{0.58} & \multicolumn{1}{c|}{0.54} & 0.54 \\ 
 & MLP & \multicolumn{1}{c|}{0.56} & \multicolumn{1}{c|}{0.49} & \multicolumn{1}{c|}{0.56} & \multicolumn{1}{c|}{0.53} & 0.54 \\ 
 & Bagging & \multicolumn{1}{c|}{0.56} & \multicolumn{1}{c|}{0.47} & \multicolumn{1}{c|}{0.58} & \multicolumn{1}{c|}{0.53} & 0.54 \\ \hline
\multirow{4}{*}{\begin{tabular}[c]{@{}l@{}}[ Cosine Similarity, \\LEAD-1,\\ Novel 1-gram \% ]\end{tabular}} & LR & \multicolumn{1}{c|}{0.61} & \multicolumn{1}{c|}{0.53} & \multicolumn{1}{c|}{0.59} & \multicolumn{1}{c|}{\textbf{0.58}} & \ \textbf{0.58} \\ 
 & SVM & \multicolumn{1}{c|}{0.60} & \multicolumn{1}{c|}{0.52} & \multicolumn{1}{c|}{0.58} & \multicolumn{1}{c|}{0.57} & 0.57 \\ 
 & MLP & \multicolumn{1}{c|}{0.60} & \multicolumn{1}{c|}{\textbf{0.54}} & \multicolumn{1}{c|}{0.55} & \multicolumn{1}{c|}{0.56} & 0.56 \\ 
 & Bagging & \multicolumn{1}{c|}{0.60} & \multicolumn{1}{c|}{0.51} & \multicolumn{1}{c|}{0.59} & \multicolumn{1}{c|}{0.56} & 0.57 \\ \hline
\multirow{4}{*}{\begin{tabular}[c]{@{}l@{}}[ Cosine Similarity,\\ LEAD-1, EXT-ORACLE\\ Novel 1-gram \%,\\ Novel 2-gram \% ]\end{tabular}} & LR & \multicolumn{1}{c|}{\textbf{0.62}} & \multicolumn{1}{c|}{0.53} & \multicolumn{1}{c|}{0.59} & \multicolumn{1}{c|} {\textbf{0.58}} & \textbf{0.58} \\ 
 & SVM & \multicolumn{1}{c|}{0.60} & \multicolumn{1}{c|}{0.52} & \multicolumn{1}{c|}{0.58} & \multicolumn{1}{c|}{0.57} & 0.57 \\ 
 & MLP & \multicolumn{1}{c|}{0.60} & \multicolumn{1}{c|}{0.50} & \multicolumn{1}{c|}{\textbf{0.61}} & \multicolumn{1}{c|}{0.56} & 0.57 \\ 
 & Bagging & \multicolumn{1}{c|}{0.60} & \multicolumn{1}{c|}{0.51} & \multicolumn{1}{c|}{0.58} & \multicolumn{1}{c|}{0.56} & 0.56 \\ \hline
\end{tabular}
\caption{Headline Classification: ML baseline model results}
\label{tab:ml_baseline_results}
\end{table*}


\begin{table*}[t]
\centering
\begin{tabular}{c|ccccc}
\hline
\multirow{2}{*}{\textbf{\begin{tabular}[c]{@{}c@{}}Pre-trained \\ Model\end{tabular}}} & \multicolumn{5}{c}{\textbf{F1 Score}} \\ \cline{2-6} 
 & \multicolumn{1}{c|}{HREL} & \multicolumn{1}{c|}{MREL} & \multicolumn{1}{c|}{LREL} & \multicolumn{1}{c|}{\begin{tabular}[c]{@{}c@{}}Overall \\ (Weighted)\end{tabular}} & \begin{tabular}[c]{@{}c@{}}Overall \\ (Macro)\end{tabular} \\ \hline
IndicBERT & \multicolumn{1}{c|}{0.66} & \multicolumn{1}{c|}{0.55} & \multicolumn{1}{c|}{\textbf{0.67}} & \multicolumn{1}{c|}{0.62} & 0.63 \\ 
mBERT & \multicolumn{1}{c|}{0.66} & \multicolumn{1}{c|}{0.50} & \multicolumn{1}{c|}{0.62} & \multicolumn{1}{c|}{0.59} & 0.59 \\ 
mDeBERTa & \multicolumn{1}{c|}{0.65} & \multicolumn{1}{c|}{\textbf{0.59}} & \multicolumn{1}{c|}{\textbf{0.67}} & \multicolumn{1}{c|}{\textbf{0.63}} & \textbf{0.64} \\ 
MuRIL & \multicolumn{1}{c|}{0.66} & \multicolumn{1}{c|}{0.55} & \multicolumn{1}{c|}{0.62} & \multicolumn{1}{c|}{0.61} & 0.61 \\ 
XLMRoBERTa & \multicolumn{1}{c|}{\textbf{0.67}} & \multicolumn{1}{c|}{0.53} & \multicolumn{1}{c|}{0.65} & \multicolumn{1}{c|}{0.61} & 0.62 \\ \hline
\end{tabular}

\caption{Headline Classification: BERT baseline model results}
\label{tab:dl_baseline_results}
\end{table*}


\begin{table*}[t]
\centering
\resizebox{\textwidth}{!}{
\begin{tabular}{cl|cccc}
\hline
\multicolumn{2}{c|}{\multirow{2}{*}{\textbf{Pre-trained model}}} & \multicolumn{4}{c}{\textbf{F1 Score}}                                                                                                     \\ \cline{3-6} 
\multicolumn{2}{c|}{}                                   & \multicolumn{1}{c|}{FME+FSE+STC}          & \multicolumn{1}{c|}{SEN+WKC+USO+MLC+CBT}          & \multicolumn{1}{c|}{Overall(Weighted)} & Overall(Macro) \\ \hline
\multicolumn{2}{c|}{IndicBERT}                          & \multicolumn{1}{c|}{0.86}          & \multicolumn{1}{c|}{0.66}          & \multicolumn{1}{c|}{0.79}              & 0.76           \\ 
\multicolumn{2}{c|}{mBERT}                              & \multicolumn{1}{c|}{0.85}          & \multicolumn{1}{c|}{0.63}          & \multicolumn{1}{c|}{0.78}              & 0.74           \\ 
\multicolumn{2}{c|}{mDeBERTa}                           & \multicolumn{1}{c|}{0.85}          & \multicolumn{1}{c|}{\textbf{0.69}} & \multicolumn{1}{c|}{\textbf{0.80}}     & \textbf{0.77}  \\ 
\multicolumn{2}{c|}{MuRIL}                              & \multicolumn{1}{c|}{0.73}          & \multicolumn{1}{c|}{0.63}          & \multicolumn{1}{c|}{0.70}              & 0.68           \\ 
\multicolumn{2}{c|}{XLMRoBERTa}                         & \multicolumn{1}{c|}{\textbf{0.86}} & \multicolumn{1}{c|}{0.68}          & \multicolumn{1}{c|}{\textbf{0.80}}     & \textbf{0.77}  \\ \hline
\end{tabular}
}
\caption{Headline Classification: BERT baseline model results for Merged fine classes}
\label{tab:dl_baseline_results_2-class}
\end{table*}

\begin{table*}[t]
\centering
\begin{tabular}{l|llllll||ll}
\hline
\multirow{2}{*}{\textbf{Fine-tuned   on}}     & \multicolumn{6}{c||}{\textbf{Tested on}}                                                                                                                   & \multicolumn{2}{c}{\textbf{Data Size}}    \\ \cline{2-9} 
                                     & \multicolumn{1}{l|}{FME}  & \multicolumn{1}{l|}{STC}  & \multicolumn{1}{l|}{FSE}  & \multicolumn{1}{l|}{WKC}  & \multicolumn{1}{l|}{SEN}  & CBT  & \multicolumn{1}{l|}{Train} & Dev  \\ \hline
No fine-tuning                       & \multicolumn{1}{l|}{0.39} & \multicolumn{1}{l|}{0.23} & \multicolumn{1}{l|}{0.25} & \multicolumn{1}{l|}{0.17} & \multicolumn{1}{l|}{0.21} & 0.15 & \multicolumn{1}{l|}{-}     & -    \\ \hline
FME                                  & \multicolumn{1}{l|}{\textbf{0.45}} & \multicolumn{1}{l|}{\textbf{0.28}} & \multicolumn{1}{l|}{\textbf{0.31}} & \multicolumn{1}{l|}{0.21} & \multicolumn{1}{l|}{0.25} & 0.17 & \multicolumn{1}{l|}{8058}  & 1007 \\ 
STC                                  & \multicolumn{1}{l|}{0.43} & \multicolumn{1}{l|}{0.27} & \multicolumn{1}{l|}{0.30} & \multicolumn{1}{l|}{0.22} & \multicolumn{1}{l|}{0.23} & 0.18 & \multicolumn{1}{l|}{3949}  & 494  \\ 
FSE                                  & \multicolumn{1}{l|}{0.41} & \multicolumn{1}{l|}{0.26} & \multicolumn{1}{l|}{0.29} & \multicolumn{1}{l|}{0.22} & \multicolumn{1}{l|}{0.23} & 0.18 & \multicolumn{1}{l|}{1416}  & 177  \\ 
WKC                                  & \multicolumn{1}{l|}{0.38} & \multicolumn{1}{l|}{0.23} & \multicolumn{1}{l|}{0.28} & \multicolumn{1}{l|}{0.20} & \multicolumn{1}{l|}{0.21} & 0.15 & \multicolumn{1}{l|}{1029}  & 129  \\ 
SEN                                  & \multicolumn{1}{l|}{0.41} & \multicolumn{1}{l|}{0.26} & \multicolumn{1}{l|}{0.29} & \multicolumn{1}{l|}{0.20} & \multicolumn{1}{l|}{0.23} & 0.18 & \multicolumn{1}{l|}{2587}  & 323  \\ 
CBT                                  & \multicolumn{1}{l|}{0.39} & \multicolumn{1}{l|}{0.24} & \multicolumn{1}{l|}{0.27} & \multicolumn{1}{l|}{0.21} & \multicolumn{1}{l|}{0.22} & 0.16 & \multicolumn{1}{l|}{1501}  & 188  \\ \hline

Total (6-class)                      & \multicolumn{1}{l|}{0.43} & \multicolumn{1}{l|}{0.27} & \multicolumn{1}{l|}{0.30} & \multicolumn{1}{l|}{0.22} & \multicolumn{1}{l|}{0.25} & 0.18 & \multicolumn{1}{l|}{18540} & 2318 \\ \hline
3-class(FME,STC,FSE)                 & \multicolumn{1}{l|}{0.44} & \multicolumn{1}{l|}{\textbf{0.28}} & \multicolumn{1}{l|}{0.30} & \multicolumn{1}{l|}{0.20} & \multicolumn{1}{l|}{0.25} & 0.20 & \multicolumn{1}{l|}{13423} & 1678 \\ 
3-class(WKC,SEN,CBT)                 & \multicolumn{1}{l|}{0.40} & \multicolumn{1}{l|}{0.25} & \multicolumn{1}{l|}{0.29} & \multicolumn{1}{l|}{0.19} & \multicolumn{1}{l|}{0.23} & 0.18 & \multicolumn{1}{l|}{5117}  & 640  \\ \hline
\end{tabular}

\caption{Class-based Headline Generation results. (Metric: ROUGE-L)}
\label{tab:style_hg_resluts}
\end{table*}

\textbf{BERT-based baseline models:}
Pre-trained models like BERT excel in text classification compared to classical ML models because they leverage extensive pre-training on diverse data, capturing language nuances and context.
In our work, we fine-tuned several state-of-the-art multilingual BERT-based models, equipping them with a classification head. The classification head is a feedforward neural network added on top of the BERT model, specifically trained for our classification task. We used a specific input format where the headline and news article text were concatenated, separated by a [SEP] token, and preceded by a [CLS] token. This format ensures a unified representation of both the title and text, significantly enhancing the model's ability to process and make accurate predictions. \\

We experiment with the following models by making use of the scripts \footnote{\url{https://github.com/huggingface/transformers/tree/main/examples/pytorch/text-classification}} provided by Huggingface.

\textbf{mBERT:} mBERT \citep{devlin2018bert} is a multilingual variant of the BERT model, which supports 102 different languages. For our baseline, we fine-tune the base version of mBERT having 110M parameters.

\textbf{XLM-RoBERTa:} XLM-RoBERTa \citep{conneau2019unsupervised} is a multilingual version of the RoBERTa model, and it was pre-trained on a vast 2.5TB CommonCrawl dataset, which included text from 100 languages. For our experiments, we utilized the xlm-roberta-base variant, boasting 270 million parameters.

\textbf{MuRIL:} MuRIL \citep{khanuja2021muril} is pre-trained on 17 Indian languages, utilizing a range of datasets, including Wikipedia, CommonCrawl, PMINDIA, and Dakshina Corpora. We employed the muril-base-cased variant with 236 million parameters for our task.

\textbf{IndicBERT:} IndicBERT \citep{doddapaneni2023towards} is a multilingual BERT model trained with the Masked Language Modeling (MLM) objective on the IndicCorp v2 dataset. This model supports 23 Indic languages as well as English and boasts 278 million parameters. We used the IndicBERTv2-MLM-only version in our experiments.

\textbf{mDeBERTaV3:} mDeBERTaV3 \citep{he2021debertav3} is a multilingual adaptation of the DeBERTa model, pre-trained on a substantial 2.5TB dataset known as CC100, featuring text from 100 languages. We used the base variant of mDeBERTaV3 in our experiments.
\\

\textbf{Hyperparameters:} For all these models, we set the maximum input sequence length to 512 subword tokens, and use a batch size of 8. We use categorical cross-entropy loss with Adam optimizer and a learning rate of 2e-05. To prevent overfitting, we use early stopping criteria to stop training when the validation loss stops improving (or begins to worsen) over two consecutive epochs. All these experiments were performed using 4 GPUs (each with a VRAM of 12GB), and 30 CPUs. The results of these experiments are presented in Table \ref{tab:dl_baseline_results}.

\section{Results \& Analysis}
\label{sec:results_analysis}

From the results presented in Table \ref{tab:ml_baseline_results}, it is apparent that the integration of a feature vector in conjunction with TF-IDF encoding, featuring elements such as cosine similarity, LEAD-1, EXT-ORACLE, Novel 1-gram \%, and 2-gram \%, clearly underscores the vital role played by these features in enhancing the performance of our models when compared to models that did not employ a feature vector. Notably, the Logistic Regression (LR) model utilizing these features achieved F1 weighted and macro scores of 0.58, which represents a 3\% improvement when compared to the model that did not utilize a feature vector. \\

Furthermore, the results presented in Table \ref{tab:dl_baseline_results} underscore the superiority of state-of-the-art BERT-based models in comparison to classical machine learning models. The best model, mDeBERTa, achieved an impressive overall F1 weighted score of 0.63 and an F1 macro score of 0.64. These scores reflect a substantial 5\% improvement in F1 weighted and a 6\% improvement in F1 macro scores when compared to the best-performing feature-based ML model. \\

\begin{figure}[]
    \centering
    \includegraphics[scale=0.50]{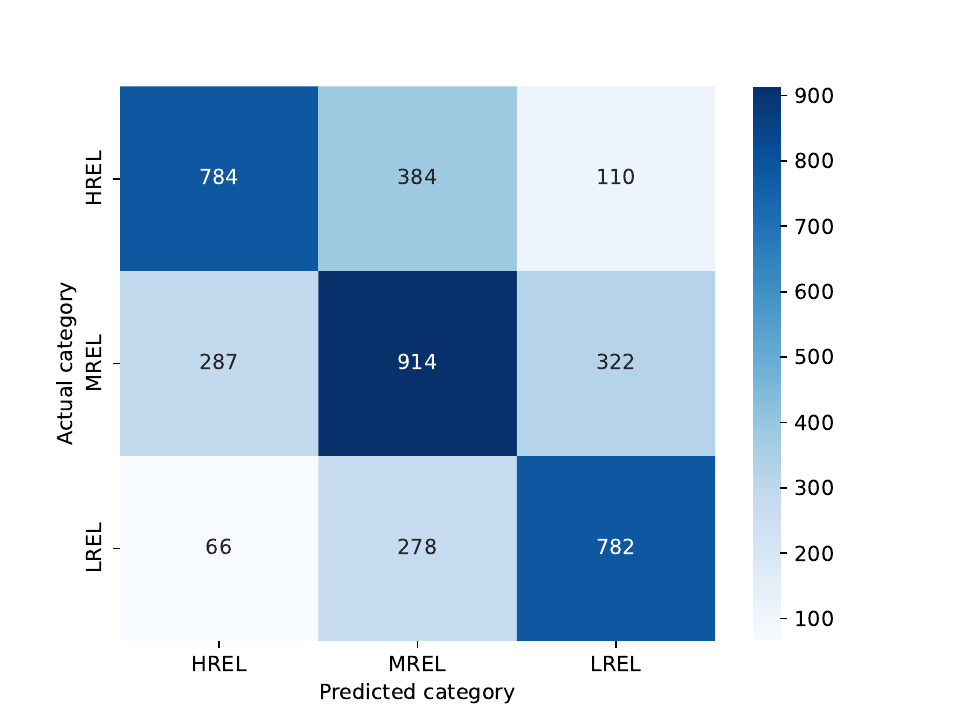}
    \caption{Confusion matrix between actual and predicted categories of mDeBERTa model}
    \label{fig:confusion_mDEBERTa_model}
\end{figure}

The confusion matrix between actual categories and predicted categories of the mDeBERTa model shown in Figure \ref{fig:confusion_mDEBERTa_model} offers valuable insights into the challenges encountered by our model. Specifically, the number of misclassifications between the Highly Related (HREL) and Moderately Related (MREL) classes highlights a notable difficulty: our model struggles to effectively distinguish between these classes. But, if we consider Factual Main Event, Factual Secondary Event and Strong Conclusion classes as relevant to the article, we see significantly better performance for DL models as seen in Table \ref{tab:dl_baseline_results_2-class}. This underscores the inherent difficulty in differentiation between highly relevant and moderately related headlines.

\section{Headline Generation}
We experimented with headline generation by using mT5 model trained on Telugu summary generation on a large Telugu dataset (Mukhyansh \cite{madasu-etal-2023-mukhyansh}). This was further fine-tuned on different subsets of TeClass to evaluate the impact of class-specific fine-tuning on the headline generation task. As seen in Table \ref{tab:style_hg_resluts}, non-fine-tuned model performs well enough but if we want the most relevant headline generation then class-aware training always significantly improves ( 5 points) ROUGE-L score across the board. In a human evaluation conducted by two volunteers on 50 news articles, we found that 34, 1, and 3 generated headlines were marked as FME, FSE, and STC respectively.

It is interesting to note that the best performance on all the relevant classes (FME, STC, FSE) is achieved by fine-tuning either on FME class or the combination of all the relevant classes. It is also interesting to see that the performance gain is not proportional to the training data size. In fact, we see a marked decrease in performance when all of the data is used. The best performance is achieved using 43\% of the data (FME).

\section{Conclusion \& Future work}
\label{sec:conclusion}

In this work, we introduce a novel, high-quality human-annotated dataset tailored for the task of relevance-based news headline classification in a low-resource language, Telugu. Our proposed dataset comprises 26,178 article-headline pairs, meticulously annotated into three primary classes: Highly Related, Moderately Related, and Least Related. Notably, this dataset stands as the largest and most diverse of its kind, encompassing various news domains and websites. This contribution marks the first dataset of its nature specifically designed for the task of headline classification in the Telugu language.\\

In our experiments with various baseline models on this dataset, our empirical findings highlight the superior performance of BERT-based models when compared to classical machine learning models. Notably, mDeBERTa achieved an impressive F1 weighted score of 0.63 and an F1 macro score of 0.64.
We firmly believe that this dataset will serve as a valuable resource for the research community working on applications such as News Headline Classification, Fake News Classification, Misinformation Classification, and other related tasks. Furthermore, the annotation guidelines and annotation process developed for this dataset can be a valuable reference for extending this task to other languages.

Further, this classification of these headlines into relevance classes assists significantly in generation of high-quality headlines at half the compute cost (with respect to a number of samples). We hope that this work will encourage attempts to extract high-quality data for generation tasks in general.

\section{Ethics Statement}
The collected news articles are subject to the respective licenses of the original websites. These resources will be released under the Creative Commons license\footnote{\url{https://creativecommons.org/licenses/by/4.0/}}, respecting individual website policies on data distribution and public availability.

\section{Acknowledgments}
We extend our sincere gratitude to Pavan Baswani for generously providing the annotation tool, which facilitated the acquisition of high-quality annotations.


\section{Bibliographical References}\label{sec:reference}

\bibliographystyle{lrec-coling2024-natbib}
\bibliography{references}

\end{document}